\documentclass[runningheads]{llncs}

\usepackage[year=2026,ID=5777]{eccv}

\usepackage{eccvabbrv}

\usepackage{graphicx}
\usepackage{booktabs}
\usepackage{amsmath}
\usepackage{amssymb}
\usepackage{float}

\usepackage[accsupp]{axessibility}

\makeatletter
\providecommand{\subparagraph}{}
\renewcommand{\subparagraph}{\@startsection
  {subparagraph}{5}{\z@}{-3.25ex\@plus -1ex \@minus -.2ex}{-1em}%
  {\normalfont\normalsize\bfseries}}
\makeatother
\usepackage{titlesec}
\titlespacing*{\section}{0pt}{8pt plus 2pt minus 2pt}{4pt plus 1pt minus 1pt}
\titlespacing*{\subsection}{0pt}{6pt plus 2pt minus 2pt}{3pt plus 1pt minus 1pt}
\titlespacing*{\paragraph}{0pt}{4pt plus 1pt minus 1pt}{0.5em}

\usepackage{caption}
\newcommand{\im}[1]{\mathbf{I}_{#1}}
\newcommand{\bg}[1]{\mathbf{B}_{#1}}
\newcommand{\fg}[1]{\mathbf{F}_{#1}}
\newcommand{\fgmotion}[2]{M_{#1\to#2}^F}
\newcommand{\bgmotion}[2]{M_{#1\to#2}^B}

\newcommand{\trimap}[1]{\mathbf{Tr}_{#1}}

\newcommand{\tianfan}[1]{}
\newcommand{\zt}[1]{}
\newcommand{\lj}[1]{}
\newcommand{\jc}[1]{}

\newcommand{\TODO}[1]{}

\usepackage[breaklinks,colorlinks,citecolor=eccvblue]{hyperref}

\begin{document}

\title{Parallax Portrait Matting\thanks{Project page: \url{https://caixin98.github.io/parallax/}}}

\titlerunning{Parallax Portrait Matting}

\author{Xin Cai\inst{1,3} \and
Jiawen Chen\inst{2} \and
Lars Jebe\inst{2} \and
Tianfan Xue\inst{1,3,4} \and
Zhoutong Zhang\inst{2}$^{\dagger}$}

\authorrunning{X.~Cai et al.}

\institute{Multimedia Laboratory, The Chinese University of Hong Kong, Hong Kong SAR, China \and
Adobe NextCam, San Jose, CA, USA \and
Shanghai AI Laboratory, Shanghai, China \and
CPII under InnoHK, Hong Kong SAR, China}

\maketitle

\begin{abstract}
Image matting is highly ill-posed, especially when both the foreground and background are richly textured.
While single-image matting methods learn strong priors from data, they often struggle on these challenging cases.
Existing approaches improve results by requiring additional signals such as green screens, polarized lighting, or clean background images, but these typically rely on specialized capture setups.
We present \emph{Parallax Portrait Matting}, a practical two-frame matting method that uses a second image captured with slight viewpoint change.
Such a setting arises naturally in burst photography, where small camera motion induces foreground-background parallax and provides complementary observations for matting.
Our pipeline estimates trimaps and foreground/background motion, then constructs aligned views for prediction.
To handle imperfect motion estimation, the network uses the background-aligned pair for direct fusion and the foreground-aligned cue through cross-attention for error compensation.
Experiments show that our method recovers finer details and more accurate foreground colors than strong single-image matting baselines on challenging portrait cases.

\end{abstract}

\begin{figure}[t]
    \centering
    \includegraphics[width=\linewidth]{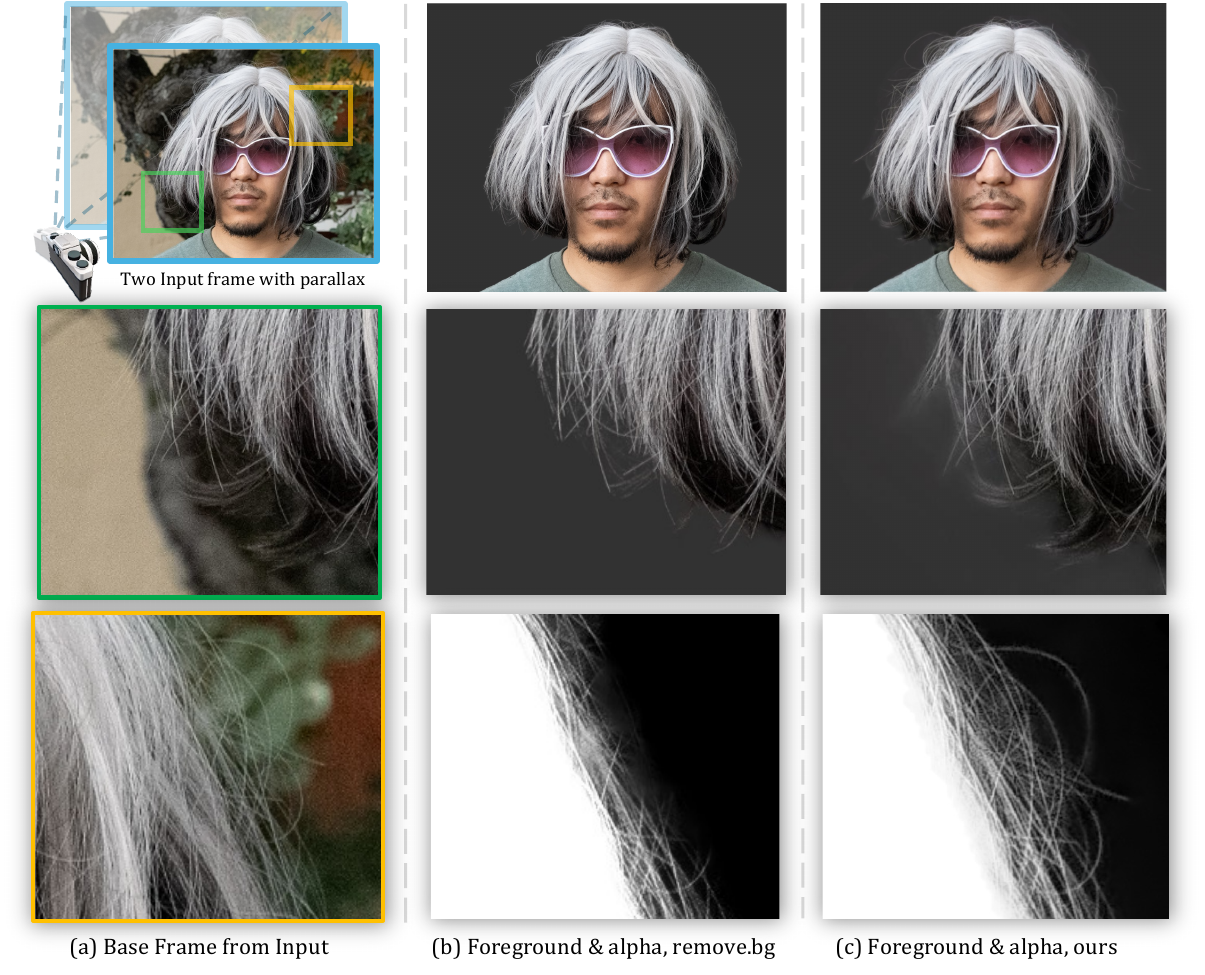}
    \caption{Our matting method exploits camera-motion-induced parallax between the foreground and the background. It takes two frames as input, each taken with a slightly different camera location, and predicts both a pre-multiplied foreground image and an alpha map. Trained on public datasets, our method produces a cleaner foreground with more details than closed-source commercial solutions like remove.bg.}
    \label{fig:teaser}
\end{figure}

\section{Introduction}
Image matting has a long history in computer vision and graphics~\cite{fry1977saga,smith_blue_screen}.
This decades-old problem aims to decompose an image $I$ into a foreground image $F$, a background image $B$, and an opacity (alpha) map $\alpha$, where they jointly reconstruct the input image through a linear composition process:
\begin{equation*}
   I = \alpha F+(1-\alpha)B. 
\end{equation*}
Like most inverse problems, the matting problem is known to be ill-posed.
To arrive at a solution representing the actual scene, it requires either priors over $F$, $B$, and $\alpha$, or additional information as constraints to the solution space.

Most modern methods resolve this ambiguity either by learning strong priors from annotated data~\cite{xu2017deep,Yu-2021-MG,li2020inductive,yao2024vitmatte,park2022matteformer} or by leveraging generative models~\cite{wang2024matting}, but single-image prediction remains difficult in highly ambiguous cases.
Another line of work tackles the ill-posedness of single-image matting by acquiring additional observations. However, existing solutions often rely on specialized capture setups such as green screens, polarization, camera arrays, focal stacks, or clean background images~\cite{aksoy2016interactive,enomoto2024polarmatte,Bando2008Extracting,joshi2006natural,joshi2007exploring,sengupta2020background}.
These approaches can produce high-quality mattes, especially for challenging regions such as hair and semi-transparent boundaries, but they typically require dedicated hardware or carefully controlled capture procedures, making them impractical for everyday photography.

In this work, we introduce \emph{Parallax Portrait Matting} (\cref{fig:teaser}), a practical two-frame matting method for portrait scenes.
Our key idea is to exploit an additional image captured with slight viewpoint change, which is often already available in casual photography or mobile burst capture.
Because the portrait subject is typically closer to the camera than the background, the two layers exhibit different apparent motion, and this parallax provides extra constraints for separating fine foreground structures from complex backgrounds.
This makes the additional frame a practical and informative cue for portrait matting without requiring specialized hardware or controlled capture setups.

The central challenge is that this cue is only useful if motion can be modeled robustly.
Off-the-shelf motion estimators~\cite{teed2020raft,xu2022gmflow} are unreliable precisely where matting is difficult: pixels that mix foreground and background do not follow a single motion.
We therefore approximate the motion in such regions as a combination of two locally smooth fields, one for the foreground and one for the background, extrapolated from nearby certain regions.
This approximation is only imperfectly valid in real capture, where wind, hair motion, or slight subject movement may introduce residual motion.
Our method is thus designed to use motion conservatively: it exploits parallax when helpful, and degrades gracefully to strong single-image predictions when motion estimation is imperfect.

Our key observation is that foreground and background motion are not equally reliable.
In typical burst portrait capture, background motion is easier to estimate: the background is often farther from the camera, its apparent motion is smoother, and it can frequently be aligned accurately enough for direct pixel-level fusion.
Foreground motion, in contrast, is much harder to estimate reliably, especially around thin structures such as hair, where local non-rigid motion and residual flow errors are common.
This asymmetry motivates our network design.
We use the background-aligned pair as the primary pixel-aligned input for matte prediction, while treating the foreground-aligned pair as a noisier auxiliary cue.
Rather than fusing it directly in pixel space, we inject it through cross-attention in feature space to compensate for residual foreground alignment errors.
This allows the model to use multi-view information when it is helpful, while remaining robust when foreground alignment is inaccurate.

We train the model on local patches and apply image and motion augmentations to improve robustness to real-world degradation and alignment noise.
Experiments show that our method outperforms single-image matting models on challenging portrait cases, producing both finer alpha mattes and substantially more accurate foreground colors than existing baselines and closed-source commercial solutions.
Accurate foreground color estimation is critical for downstream tasks such as compositing onto new backgrounds, yet most existing matting methods either ignore it entirely or produce colors contaminated by the original background.
These results suggest that a casually captured second frame is a practical and complementary cue for portrait matting.

In summary, our contributions are threefold:
\begin{itemize}
   \item We introduce \emph{Parallax Portrait Matting}, a practical two-frame portrait matting formulation that leverages an additional image with slight viewpoint change to better constrain the decomposition.
   \item We develop a robust matting framework that uses parallax cues under imperfect motion estimation, with direct fusion from reliable background alignment and feature-level compensation from noisier foreground alignment.
   \item We show empirically that a captured second frame improves both alpha matte quality and foreground color accuracy over strong single-image baselines on challenging portrait cases, while remaining robust to motion estimation errors.
\end{itemize}

\section{Related Work}

\noindent\textbf{Single Image Matting.} Most existing work aims to predict alpha and/or foreground color from a single input image.
Many methods use additional guidance signals such as semantic segmentation~\cite{Yu-2021-MG, li2020inductive, yaman2022alpha}, instance segmentation~\cite{huynh2024maggie,sun2022human}, a trimap~\cite{he2011global, park2022matteformer, tang2019learning, hou2019context} or different types of user annotations~\cite{li2024matting, liu2020boosting}.
More recent learning-based methods take as input a single image without any guidance, either by implicitly incorporating guidance signal prediction~\cite{Ke-2022-MODNet,shen2016deep, chen2018semantic, yao2024vitmatte, li2020natural}, or by leveraging strong learned priors, \eg from generative models~\cite{wang2024matting}.
Such generative priors have also proven effective for recovering or preserving fine high-frequency details in other ill-posed reconstruction problems, such as lensless image reconstruction~\cite{cai2024phocolens} and high-compression latent diffusion~\cite{cai2026davae}.
Since the matting problem is intrinsically ill-posed, those methods usually fail to recover fine details if the background is highly textured.

\noindent\textbf{Video Matting.}
Video matting methods aim to predict alpha and/or decontaminated color for an entire sequence. Most learning-based methods extend beyond current single-image matting models with temporal feature aggregation designs, such as graph neural networks~\cite{wang2021video}, temporal RNNs~\cite{rvm}, transformers~\cite{li2024vmformer}, deformable convolution~\cite{sun2021deep}, or temporal image difference~\cite{sun2023ultrahigh}. All these methods aim to learn such feature aggregation end-to-end with video data supervision, making single-image matting consistent over the entire video sequence. The most similar work to ours is~\cite{chuang2002video}, where the authors use optical flow to correlate frames in the video to better estimate the background, then optimize for per-frame alpha maps. Our method differs from video matting methods in that we focus on utilizing motion between frames to improve single-image matting results for challenging cases.

\noindent\textbf{Matting with additional signals.} Additional information helps better condition the matting problem. Background matting~\cite{sengupta2020background, lin2021real} uses an additional background image as conditioning and is able to produce high quality matting for images and videos. Polarization systems~\cite{enomoto2024polarmatte} are able to capture ground truth transmittance maps. Attaching a color filter~\cite{Bando2008Extracting} to a camera lens allows one to simultaneously capture multiple views from a single exposure, where stereo algorithms can provide richer details than traditional matting models. One can also achieve the same effect with a camera array~\cite{joshi2006natural}. In addition, focal stacks~\cite{joshi2007exploring} can be used to provide additional information since foreground and background are blurred differently. Our method differs from those methods in that we do not ask the user to perform additional setups before capturing; we simply require an additional image.
\section{Formulation \& Assumptions} \label{sec:problem_formulation}
We first describe the problem formulation and examine how the parallax between the foreground and background can better condition the matting equation.
We then introduce the assumptions made by our formulation and discuss how realistic they are. Finally, we discuss the building blocks of our matting pipeline which includes foreground and background motion estimation, trimap generation, and the design of our matting prediction network.

For a single image $\im{1}$, the matting problem tries to decompose it into a foreground image $\fg{1}$, a background image $\bg{1}$ and an alpha map $\alpha_1$ that encodes the opacity of the foreground image, where:
\begin{equation*}
    \im{1} = \alpha_1\fg{1} + (1-\alpha_1)\bg{1}.
\end{equation*}
This linear system is underdetermined, with only one constraint but three unknowns (in grayscale; in color it has three constraints and seven unknowns). Unless we impose priors over each of the unknowns, there are an infinite number of solutions that all satisfy the equation.

Now assume that we capture another frame $\im{0}$ which differs from $\im{1}$ due to parallax between the foreground object and the background.
Let us describe this parallax with two motion fields $\fgmotion{1}{0}(\cdot)$ and $\bgmotion{1}{0}(\cdot)$, where $\fgmotion{1}{0}(\cdot)$ is the warping function for the foreground motion and $\bgmotion{1}{0}(\cdot)$ is the background motion. The matting equation for $\im{0}$ is:
\begin{equation*}
\im{0} = \alpha_0\fg{0} + (1-\alpha_0)\bg{0}.
\end{equation*}
We can correlate $\alpha_0$, $\fg{0}$ and $\bg{0}$ with $\alpha_1$, $\fg{1}$ and $\bg{1}$ by:
\begin{equation*}
\alpha_0 = \fgmotion{1}{0}(\alpha_1), \fg{0} = \fgmotion{1}{0}(\fg{1}), \bg{0} = \bgmotion{1}{0}(\bg{1}).
\end{equation*}
Substituting into the equation above, we have:
\begin{equation*}
\im{0} = \fgmotion{1}{0}(\alpha_1 \fg{1}) + \bigl(1-\fgmotion{1}{0}(\alpha_1)\bigr)\bgmotion{1}{0}(\bg{1}).
\end{equation*}
The objective of parallax portrait matting is to solve for the foreground $\fg{1}$ and alpha map $\alpha_1$ of the base frame, given the additional information provided by the alternate frame $\im{0}$.

The key idea for parallax matting is that, given a correct motion estimate ($\fgmotion{1}{0}$ and $\bgmotion{1}{0}$), the extra frame we observe serves as another constraint on the \emph{same} unknowns we want to estimate.
Note that the constraint is only useful if the parallax between foreground and background exists.
At pixels where $\fgmotion{1}{0}$ is the same as $\bgmotion{1}{0}$, the equations are linearly dependent.
This assumption in turn imposes some mild conditions on the scene and capture process.

\noindent\textbf{Parallax between foreground and background.} We aim to extract the foreground, which is by definition closer to the camera than the background. In other words, there should be sufficient motion parallax between the two layers.

\noindent\textbf{Mostly static scene.} The second assumption is that the scene is almost static, so that we can model parallax simply through two warping fields, one for the foreground and one for the background. Our method robustly handles this assumption by being tolerant of motion estimation errors.

\noindent\textbf{Consistent camera settings.} Finally, we assume that both frames are captured using the same settings (exposure, white balance, color, tone, etc).
Most cameras feature an auto-exposure-and-lock function, which we use when capturing our dataset.
In our experiments, we capture raw images and render them with the same parameters in Adobe Lightroom to ensure maximum consistency.

Together, these conditions define the operating regime of our method: small-baseline burst portraits with visible foreground-background parallax, a mostly static subject and scene, and reasonably consistent appearance across the two frames.
We emphasize that they are not hard requirements. Because our design uses motion conservatively---trusting the reliable background alignment for direct fusion while treating foreground alignment only as a noisy feature-level cue---the model degrades gracefully toward single-frame behavior when parallax is weak or alignment is unreliable, rather than failing catastrophically.

\section{Method} \label{sec:method}

\begin{figure}[htbp]
\centering
\includegraphics[width=\linewidth]{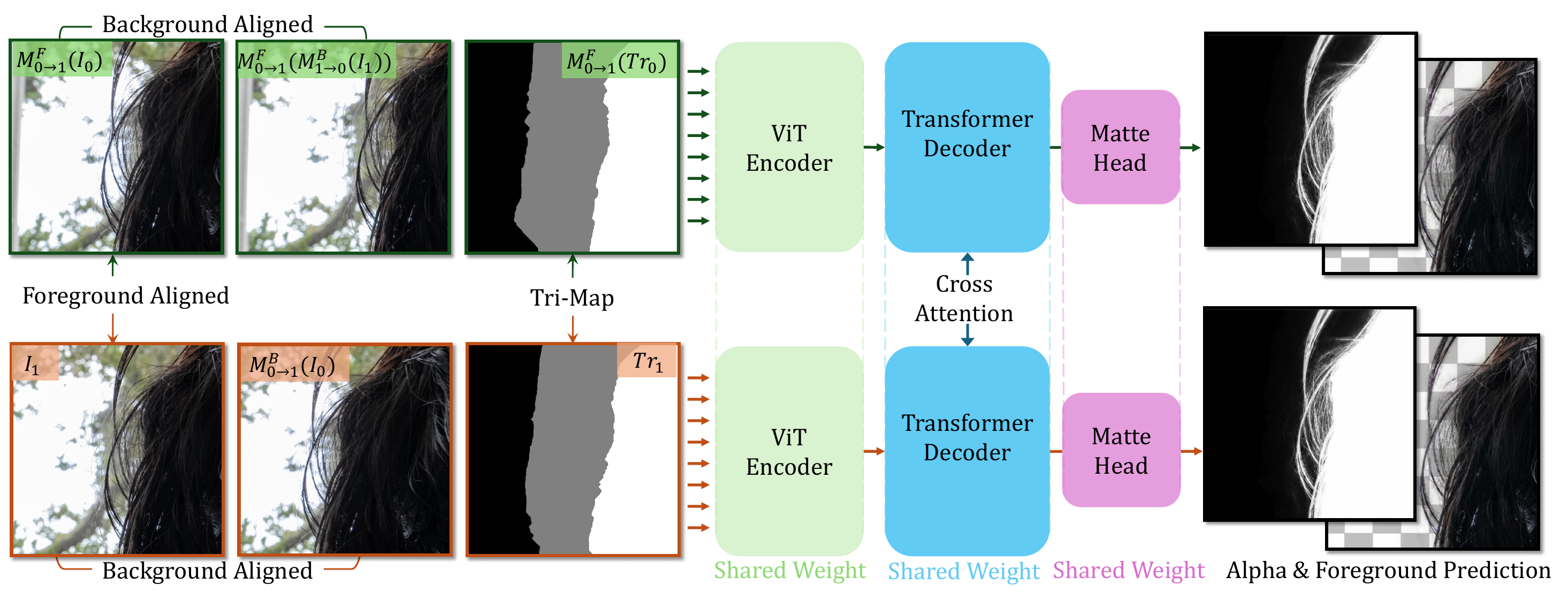}
\caption{
Overview of our framework.
The bottom branch is the primary prediction path and uses the base frame together with a background-aligned alternate frame, which provides reliable pixel-aligned evidence for matting.
The top branch uses a foreground-aligned view as an auxiliary cue.
Because foreground alignment is often imperfect, information from this branch is incorporated through cross-attention rather than direct pixel-space fusion, allowing the network to compensate for foreground motion errors while suppressing unreliable correspondences.
}\label{fig:framework}
\end{figure}

Given two portrait images captured from slightly different viewpoints, our goal is to predict the foreground and alpha matte of a chosen base frame.
Our pipeline (\cref{fig:framework}) first estimates a trimap for each frame, then computes foreground and background motion fields, and finally uses the resulting aligned views as input to a matting network.
The network takes both background-aligned and foreground-aligned observations together with the trimaps, and predicts the foreground and alpha map of the base frame.

In practice, background alignment is usually more reliable than foreground alignment.
We therefore use the background-aligned pair as the primary input for direct fusion, and use the foreground-aligned cue only through cross-attention in feature space to compensate for residual foreground motion errors.
This design allows the model to exploit parallax when it is informative, while remaining robust when foreground alignment is imperfect.

\subsection{Trimap Estimation} \label{sec:trimap}

Following most prior work in image matting, we first create trimaps.
Starting with a state-of-the-art dichotomous segmentation network, BiRefNet~\cite{zheng2024bilateral}, we generate a binary foreground mask and then erode and dilate it by 100 pixels to produce foreground and background regions, respectively (see \cref{fig:motion-estimation}). The resulting uncertain band is 200 pixels wide, which is generous enough to cover the error margin of modern segmentation models.

\subsection{Motion Estimation} \label{sec:alignment}
\begin{figure}[htbp]
  \centering
  \includegraphics[width=\linewidth]{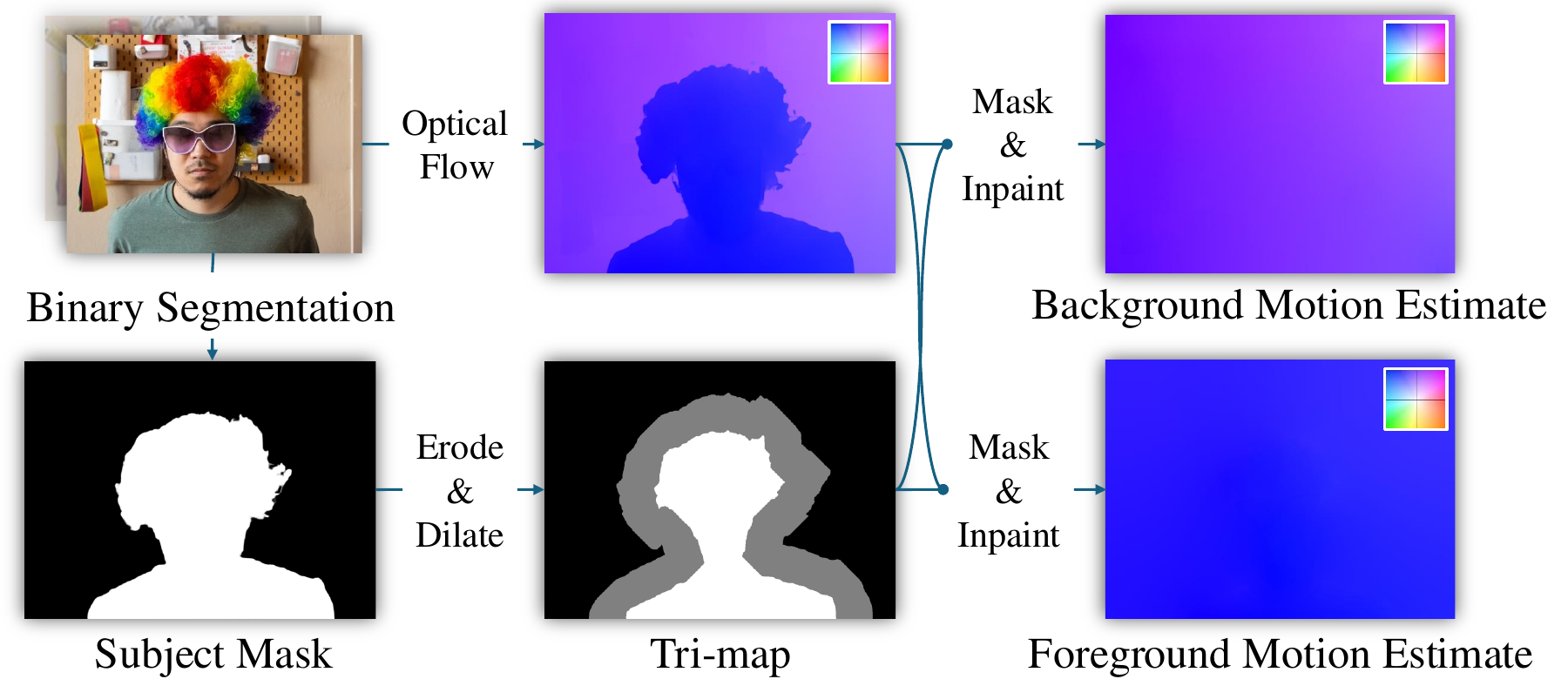}
  \caption{Our motion estimation pipeline. Given two frames, we first estimate optical flow between the two, together with a trimap generated as described in~\cref{sec:trimap}. Since flow estimation in overlapping areas is often incorrect, we inpaint them by their nearest neighbor in non-overlapping regions.}
  \label{fig:motion-estimation}
\end{figure}

Given the trimap, we proceed to estimate the motion fields $\bgmotion{0}{1}$ and $\fgmotion{0}{1}$. Due to complex occlusions between foreground and background in the uncertain region, separately estimating motion for foreground and background is very hard. To circumvent this challenge, we follow the common assumption~\cite{bhat2021deep} that motion for both the foreground and the background is locally smooth, and therefore motion in the uncertain region can be estimated by extrapolating motion estimates from certain regions. Specifically, we first estimate optical flow between two images with an off-the-shelf method such as GMFlow~\cite{xu2022gmflow}. To handle occlusion, we simply replace the estimated motion in uncertain regions with the values of its nearest neighbor inside the certain region, both for the foreground and background. \cref{fig:motion-estimation} shows an example of this motion estimation process. 
\begin{figure}[htbp]
  \centering
  \includegraphics[width=0.8\linewidth]{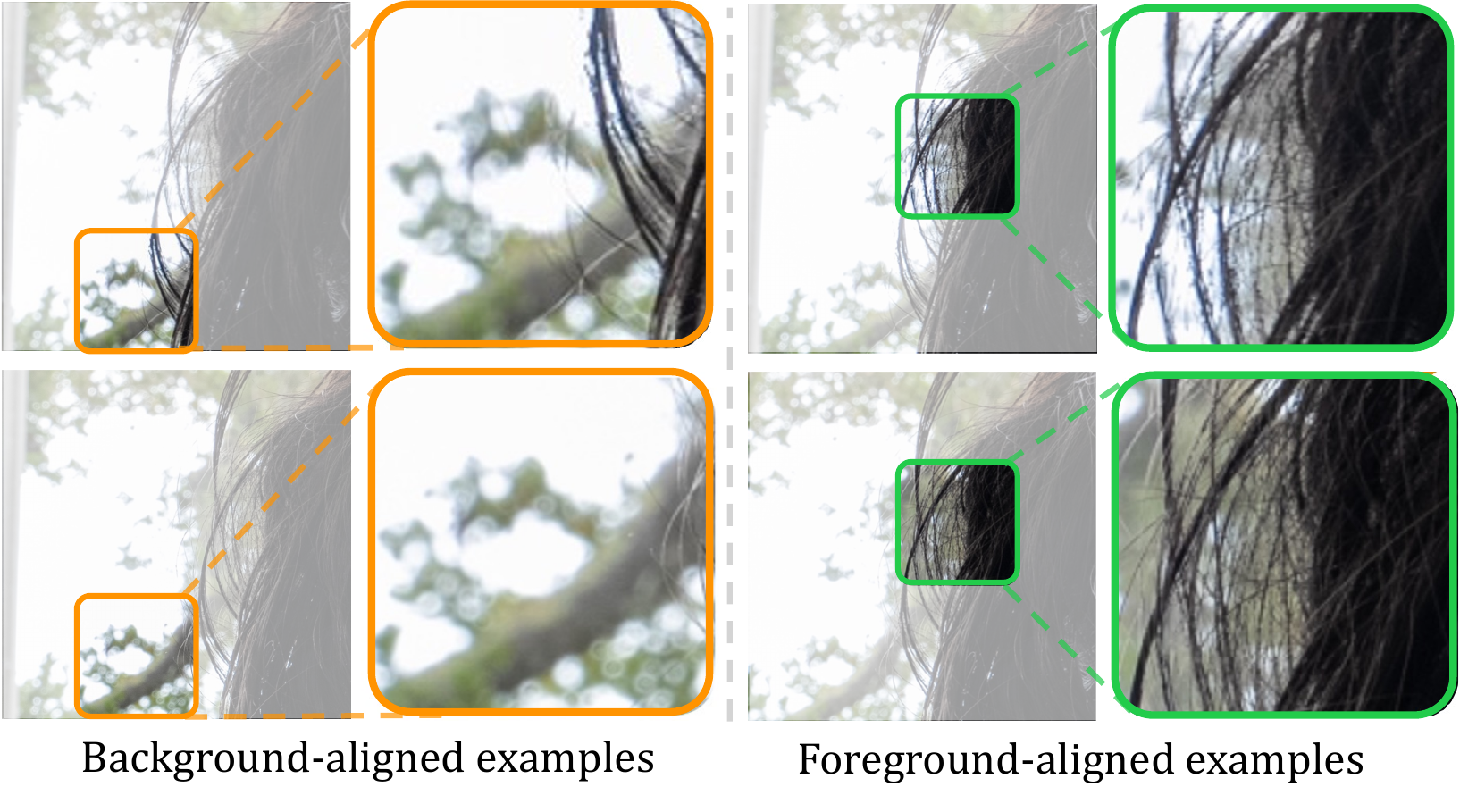}
  \caption{An example of foreground and background alignment. Background-aligned patches only contain foreground subject motion, and the dis-occluded background in the alternate frame provides more context. Foreground-aligned patches fixate on the subject, which directly helps in separating the foreground.}
  \label{fig:alignment-example}
\end{figure}

With these motion estimates, we can more intuitively see how they are helpful for the matting problem. For example, if we warp the image $\im{0}$ with the background motion $\bgmotion{0}{1}$, we get an image $\im{0\to1}^B$ that differs only from $\im{1}$ in foreground regions. Regions that are originally occluded might become dis-occluded and therefore provide a strong hint on what the occluded background is. If we warp $\im{0}$ using $\fgmotion{0}{1}$, then we get an image $\im{0\to1}^F$ where the foreground stays put and the background has shifted, providing a strong signal on what the foreground object is. \cref{fig:alignment-example} shows an illustration of this intuitive result. Therefore, we would like the network to utilize such motion information by looking at warped frames using both the foreground flow and the background flow. However, reliably and robustly doing so requires a specialized design.

\subsection{Foreground and Alpha Estimation}
\label{sec:patch-level-matting}

A straightforward design is to warp the alternate frame using both estimated motion fields and directly concatenate all aligned images in pixel space to predict the alpha matte and foreground color.
In practice, however, this strategy is brittle because it implicitly assumes that foreground and background alignments are equally reliable.
This assumption does not hold in real captures.
Background motion is usually easier to estimate and often remains smooth across the image, so the background-aligned image $\im{0\to1}^B=\bgmotion{0}{1}(\im{0})$ typically provides relatively trustworthy pixel correspondences with the base image $\im{1}$.
Foreground motion, by contrast, is much harder to estimate accurately due to local non-rigid motion, especially around thin structures such as hair.
As a result, directly concatenating a foreground-aligned image $\im{0\to1}^F=\fgmotion{0}{1}(\im{0})$ may inject incorrect pixel-level evidence and confuse the prediction network.

To address this issue, we explicitly decouple \emph{reliable fusion} from \emph{error compensation}.
Our model uses the background-aligned pair as the primary input for matte prediction, since $\im{1}$ and $\im{0\to1}^B$ are already approximately aligned in pixel space and can therefore be fused directly.
Concretely, one branch of our framework takes the base image $\im{1}$, the background-aligned image $\im{0\to1}^B$, and the trimap $\trimap{1}$ as input, and predicts the foreground image $F_1$ and alpha map $\alpha_1$ of the base frame.
This branch serves as the main prediction path and is responsible for learning a strong and stable matting solution.

We then introduce a second branch based on foreground alignment.
Instead of treating the foreground-aligned image as another source of hard pixel-aligned evidence, we use it as a noisy auxiliary cue.
Specifically, the second branch takes the foreground-aligned image $\im{0\to1}^F=\fgmotion{0}{1}(\im{0})$, a companion frame $\fgmotion{0}{1}(\bgmotion{1}{0}(\im{1}))$, and the warped trimap $\fgmotion{0}{1}(\trimap{0})$ as input.
This branch predicts the alpha and foreground of the foreground-aligned view.
During decoding, the two branches interact through cross-attention layers, which allow the model to establish soft correspondences in feature space rather than relying on exact pixel alignment.
In this way, the network can selectively borrow useful information from the foreground-aligned branch where the correspondence is reliable, while suppressing regions affected by motion estimation errors.

Our framework is therefore symmetric in architecture but asymmetric in function:
the background-aligned stream provides reliable pixel-level evidence for direct fusion, whereas the foreground-aligned stream provides feature-level correction for residual alignment errors.
The cross-attention mechanism is particularly important near ambiguous regions such as thin structures and semi-transparent boundaries, where foreground motion is most difficult to estimate accurately.
Moreover, because the main branch already forms a strong prediction path using the base image and the more reliable background-aligned observation, our model naturally remains robust when parallax is weak or foreground alignment is inaccurate.
The shared weights between the two branches further encourage a unified representation, allowing the model to benefit from both single-image matting data and multi-frame parallax cues.
As a result, the network can exploit foreground parallax when it is informative, while gracefully falling back to stable single-frame behavior when the auxiliary motion cue is noisy.

\subsection{Training Objectives}

Our training loss consists of an alpha loss and a pre-multiplied foreground color loss, described below.

\noindent\textbf{Alpha Loss.} To supervise the alpha prediction, we use an $L_1$ loss. However, in a given image, only a small set of pixels have an alpha value between 0 and 1, which biases the training towards modeling pixels that have an alpha value of 0 and 1. To mitigate this, we adaptively weight those pixels by normalizing them separately. Specifically, we define
\begin{equation*}
    \mathcal{L}_{\text{sep}} =  \frac{1}{|S|} \sum_{x \in S} \left| \alpha[x] - \alpha^{\text{gt}}[x] \right| + \frac{1}{|H|} \sum_{x \in H} \left| \alpha[x] - \alpha^{\text{gt}}[x] \right|,
\end{equation*}
where $S$ denotes the set of pixels that are ``soft'', meaning they have a ground truth alpha value between 0 and 1, and $H$ denotes the set of pixels that are ``hard'' and have a ground truth alpha value of exactly 0 or 1. Following prior work~\cite{Ke-2022-MODNet,yao2024vitmatte}, we also use a Laplacian loss $\mathcal{L}_{\text{laplacian}}$ which calculates the $L_1$ loss after applying a Laplacian filter, and a gradient penalty loss $\mathcal{L}_{\text{grad}}$ which calculates an $L_1$ loss over spatial gradients of the alpha map.

\noindent\textbf{Foreground Color Loss.} To improve robustness, we ask our network to predict a pre-multiplied foreground image $(\alpha F)_i$ of the image $\im{i}$ along with the alpha map. We supervise the pre-multiplied foreground prediction with a composition loss. That is, we recompose our predicted pre-multiplied foreground color back to the original image using the ground truth alpha and the background image. Formally, the composition loss can be written as:
\begin{equation*}
    \mathcal{L}_{\text{composition}} = \left| I_i - (\alpha_i F_i + (1 - \alpha^{gt}_i) B^{gt}_i) \right|,
\end{equation*}
where $I_i$ is the original pixel value, $\alpha_i F_i$ is the predicted pre-multiplied foreground color, and $(1 - \alpha^{gt}_i) B^{gt}_i$ is the ground truth pre-multiplied background color.

Combining all the components, the total loss function is:
\begin{equation*}
    \mathcal{L}_{\text{total}} = \mathcal{L}_{\text{sep}} + \mathcal{L}_{\text{laplacian}} + \mathcal{L}_{\text{grad}} + \mathcal{L}_{\text{composition}}.
\end{equation*}
\subsection{Patch-Based Training and Inference}
{\sloppy
Since matting usually involves high-resolution inputs, we train and test our model on selected local patches that contain mixtures of foreground and background. Specifically, we first prepare full-resolution inputs, alignment results ($\im{1}$, $\bgmotion{0}{1}(\im{0})$, $\fgmotion{0}{1}(\im{0})$, and $\fgmotion{0}{1}(\bgmotion{1}{0}(\im{1}))$), and trimaps ($\trimap{1}$ and $\fgmotion{0}{1}(\trimap{0})$). We then use the trimap $\trimap{1}$ of the base frame to extract $448 \times 448$ patches covering all uncertain regions. To ensure smooth transitions when fusing patches, we leave an overlap of $224$ pixels between neighbors. After predicting the alpha map and foreground color for each patch, we merge overlapping patches with a Gaussian window function to produce the final result.\par}
\section{Experiments}

\subsection{Training and Implementation Details}

Due to the unsatisfactory ground-truth quality in real data~\cite{wang2024matting}, our model is trained only on synthetic data. We create synthetic training samples by randomly compositing foreground subjects onto background images while simulating camera-induced parallax. The foreground subjects are sourced from two real-world portrait datasets—P3M-10K~\cite{li2021privacy} and HHM-2K~\cite{sun2023ultrahigh}—which contain 9,421 and 2,000 high-resolution images, respectively. Although these datasets provide imperfect alpha mattes and lack foreground color annotations, we generate pseudo-foreground color annotations using the layer-diffusion strategy~\cite{zhang2024transparent}. The background images are drawn from BG-20K~\cite{li2022bridging}, which supplies 15,000 images for training. During composition, we further augment the alpha matte annotations through random gamma transformations. To reduce the gap between real and synthetic data and increase training difficulty, we also apply histogram equalization to 50\% of the foreground images, aligning their color distribution with that of the background.

To simulate motion, we apply random affine transformations to both the foreground subject and background image before composition, following~\cite{rvm}. Because our network processes warped images, we add random noise (approximately 10 pixels) to these transformations so that the network never encounters perfectly aligned patches during training.

Given that our method relies on a rough trimap estimate as input, we first binarize the ground-truth alpha mask during training and then generate a trimap by applying random dilation/erosion operations (typically conducted for 60 to 120 iterations). At inference time, we generate the trimap using the strategy described in~\cref{sec:trimap}.

Our matting network uses two ViT-Small models~\cite{dosovitskiy2020image} for the encoder and decoder, and a ViTMatte head~\cite{yao2024vitmatte}, totaling 38.6M parameters, which is comparable to or fewer than MatteFormer (44.7M) and MatAnyone (35.2M). The network is initialized with weights from the CroCo pre-trained model~\cite{weinzaepfel_croco_nips_2022}, whose cross-view completion pre-training is a natural fit for our correspondence-aware two-frame prediction; this is an initialization choice rather than a required component of our formulation, and stronger cross-view or 3D foundation models could be substituted in future work. The full pipeline additionally uses BiRefNet~\cite{zheng2024bilateral} (200M) for trimap estimation and GMFlow~\cite{xu2022gmflow} (7.4M) for motion estimation, both of which are off-the-shelf components and can be replaced with more efficient alternatives. Training is conducted on 8 NVIDIA RTX 4090 GPUs with a batch size of 2 per GPU using the AdamW optimizer with a learning rate of $5\times10^{-5}$. We train for 50 epochs, each consisting of 100K synthetic patch pairs.

\subsection{Evaluation on Synthetic Data}

\noindent\textbf{Datasets.} Our test set is constructed similarly to our training dataset, but we use a different real-world portrait matting dataset for testing. Specifically, we use P3M-500-NP~\cite{li2021privacy}, PPM-100~\cite{Ke-2022-MODNet}, and RWP-636~\cite{Yu-2021-MG} for foreground, and we use BG-20K's test set as background.

\noindent\textbf{Metrics.} We use common evaluation metrics~\cite{rhemann2009perceptually} to measure the accuracy of our predicted alpha maps, which include the Sum of Absolute Differences (SAD), Mean Squared Error (MSE), Connectivity (Conn), and the spatial gradient (Grad) metric. 
We follow the common practice to scale up the SAD and MSE numbers by $10^3$ for better readability. To measure how accurate our foreground color estimation is, we calculate the MSE between the estimated pre-multiplied foreground colors $\alpha F$ and the ground truth. For methods that do not predict (pre-multiplied) foreground color, we follow the protocol of~\cite{rvm}: we multiply the input frame by the predicted alpha matte and treat the result as the pre-multiplied foreground color.

\noindent\textbf{Baselines.} We compare against several state-of-the-art single-image matting and video matting methods. For trimap-free methods, we compare against MODNet~\cite{Ke-2022-MODNet} and ViTAE-S~\cite{MA-2023-P3M-ViTAE}. For trimap-based methods, we compare against MG-Matting~\cite{Yu-2021-MG} and MatteFormer~\cite{park2022matteformer}. We further evaluate our method against video matting methods MaGGIe~\cite{huynh2024maggie} and MatAnyone~\cite{yang2025matanyone}. Among the baselines, ViTAE-S, MODNet, and MatteFormer only predict alpha; for these methods we use the input image masked by the predicted alpha as their foreground color.
For a fair comparison, all trimap-based methods (MG-Matting, MatteFormer, and ours) use the \emph{same} BiRefNet-generated trimap described in~\cref{sec:trimap}; no manual or method-specific trimaps are used, so the comparison does not depend on trimap quality. The trimap-free methods are included only as single-image context.
The video matting baselines are designed for full-sequence temporal aggregation rather than two-frame parallax; we therefore treat them as adjacent-frame references, and their gap to our method reflects the benefit of explicit parallax reasoning rather than mere access to additional frames.

\begin{table}[!t]
    \centering
    \caption{Quantitative comparison on synthetic test set.}
    \label{tab:benchmark-tab}
    \setlength\tabcolsep{1.5pt}
    \resizebox{\linewidth}{!}{
    \begin{tabular}{ll ccccc ccccc ccccc}
        \toprule
        & & \multicolumn{5}{c}{PPM-100} & \multicolumn{5}{c}{P3M-NP-500} & \multicolumn{5}{c}{RWP-636} \\
        \cmidrule(lr){3-7}\cmidrule(lr){8-12}\cmidrule(lr){13-17}
        Method & Input & SAD & MSE & Conn & Grad & MSE$(\alpha F)$ &
        SAD & MSE & Conn & Grad & MSE$(\alpha F)$ &
        SAD & MSE & Conn & Grad & MSE$(\alpha F)$ \\
        \midrule
        MODNet~\cite{Ke-2022-MODNet} & TF & 28.02 & 36.81 & 11.27 & 16.80 & 7.45 & 34.29 & 68.64 & 14.11 & 22.17 & 14.96 & 60.46 & 119.25 & 36.28 & 60.36 & 28.97 \\
        ViTAE-S~\cite{MA-2023-P3M-ViTAE} & TF & 18.24 & 25.02 & 8.35 & 15.26 & 6.16 & 14.82 & 26.02 & 7.81 & 13.14 & 6.95 & 24.30 & 37.70 & 17.94 & 39.92 & 10.26 \\
        MG-Matting~\cite{Yu-2021-MG} & TB & 49.99 & 73.76 & 31.95 & 22.34 & 15.97 & 35.64 & 53.01 & 22.41 & 15.93 & 12.86 & 49.42 & 85.92 & 29.27 & 32.67 & 20.81 \\
        MatteFormer~\cite{park2022matteformer} & TB & \underline{5.53} & \underline{2.72} & \underline{3.54} & \underline{3.46} & \underline{0.96} & \underline{4.90} & \underline{2.99} & \underline{2.93} & \underline{3.40} & \underline{1.08} & \underline{8.43} & \underline{7.14} & \underline{5.82} & \underline{7.46} & \underline{2.11} \\
        MaGGIe~\cite{huynh2024maggie} & Video & 27.41 & 7.86 & 5.51 & 9.72 & 2.04 & 18.77 & 7.02 & 6.04 & 6.35 & 2.47 & 21.33 & 15.48 & 10.64 & 14.92 & 4.79 \\
        MatAnyone~\cite{yang2025matanyone} & Video & 26.88 & 8.07 & 5.64 & 7.83 & 2.13 & 18.54 & 6.94 & 5.81 & 6.12 & 2.33 & 17.65 & 13.40 & 8.79 & 12.42 & 4.63 \\
        Ours & 2-view & \textbf{4.13} & \textbf{2.28} & \textbf{3.26} & \textbf{2.98} & \textbf{0.55} & \textbf{3.45} & \textbf{1.75} & \textbf{2.13} & \textbf{2.48} & \textbf{0.59} & \textbf{5.57} & \textbf{3.51} & \textbf{4.72} & \textbf{6.74} & \textbf{1.37} \\
        \bottomrule
    \end{tabular}
    }
    {\raggedright\scriptsize TF = trimap-free, TB = trimap-based.\par}
\end{table}

\begin{figure}[htbp]
\centering
\includegraphics[width=\linewidth]{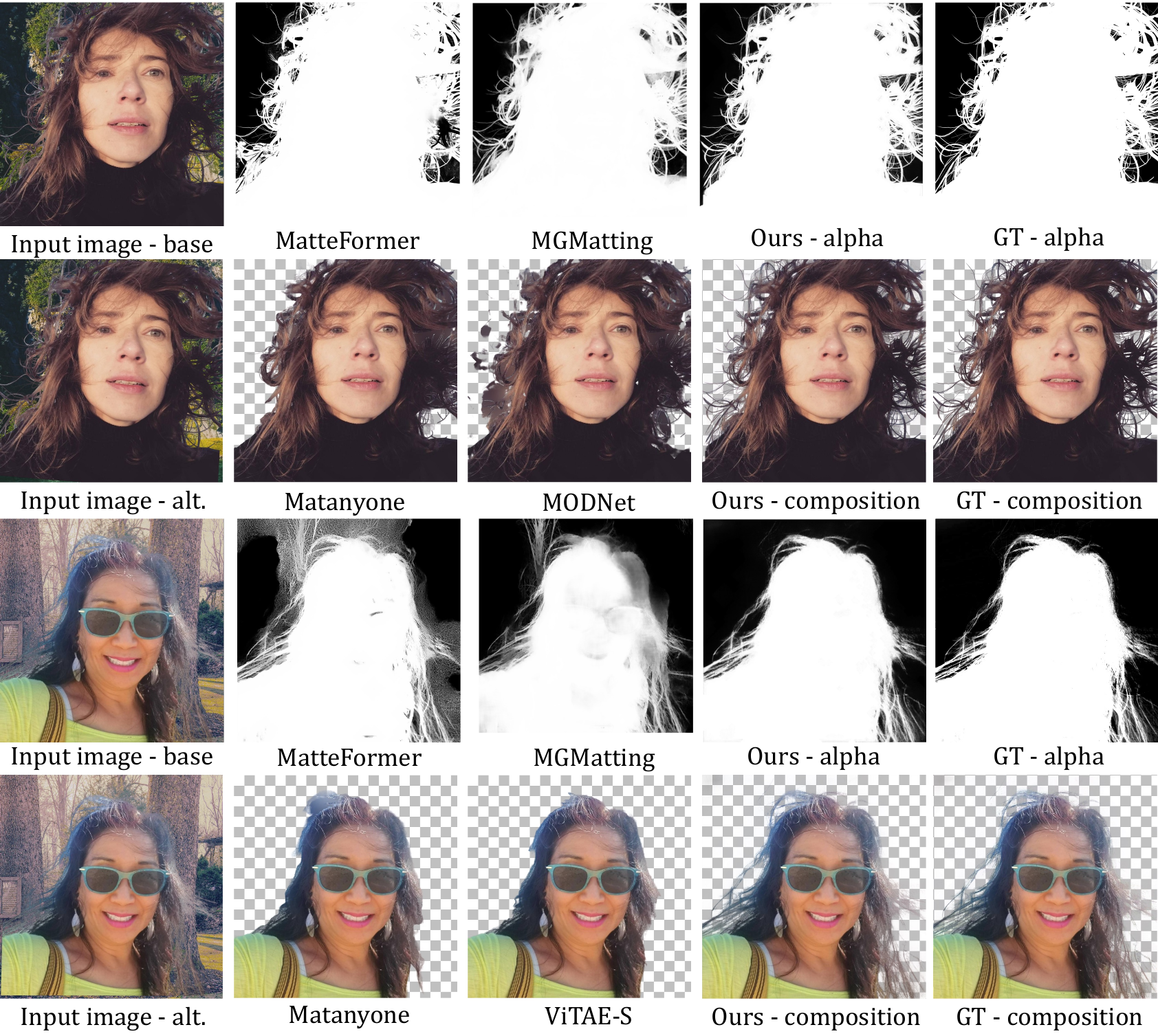}
\caption{Qualitative results on synthetic test sets.}
\label{fig:synth_result}
\end{figure}

\noindent\textbf{Quantitative Results.} As shown in~\cref{tab:benchmark-tab}, our approach consistently outperforms all baselines across every metric, validating that motion is a strong complementary signal for matting. The gain is especially pronounced in foreground color accuracy (MSE of $\alpha F$), where our method reduces error by 35--45\% over the best baseline—critical for downstream compositing.

\noindent\textbf{Qualitative Results.} \cref{fig:synth_result} shows qualitative results from all baselines, our method, and the ground-truth annotation. Note that single-image matting struggles when the background is cluttered, where one cannot reliably distinguish the foreground subject from the background using a single frame. Motion effectively separates the two, and our method can recover fine details in such cases.
\begin{figure}[htbp]
\centering
\includegraphics[width=\linewidth]{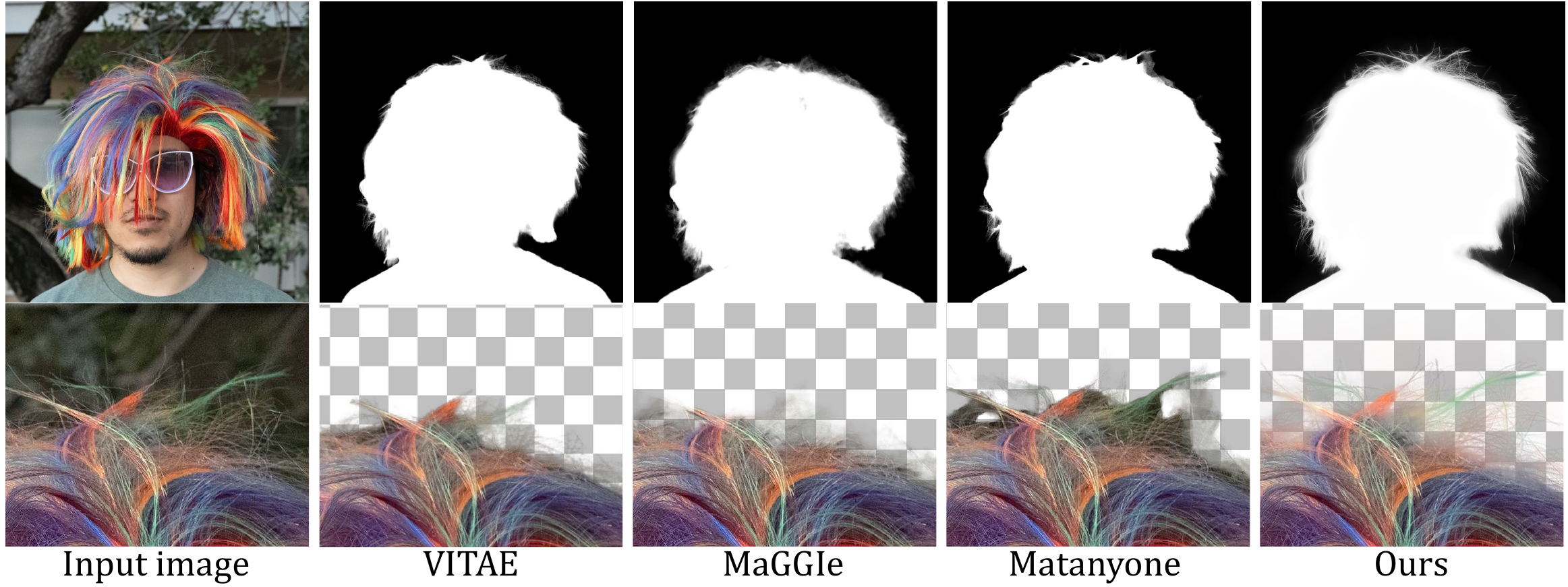}\\[2pt]
\includegraphics[width=\linewidth]{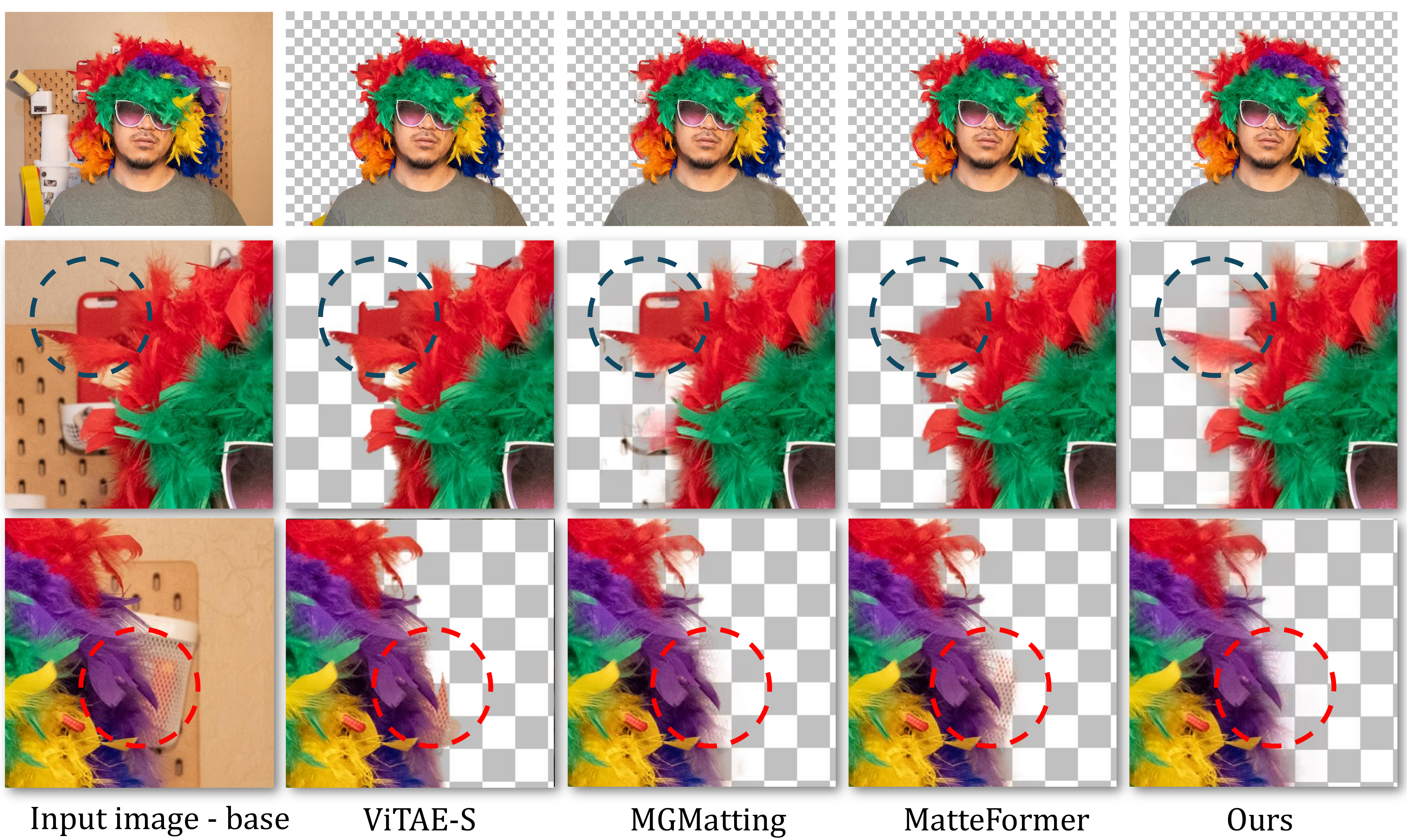}
\caption{Qualitative results on real-world images. Our method produces higher-quality alpha mattes, more accurate foreground colors, and finer structural details compared to existing methods.}
\label{fig:realworld_result}
\end{figure}

\subsection{Evaluation on Real-World Data}
\begin{figure}[htbp]
\centering
\includegraphics[width=\linewidth]{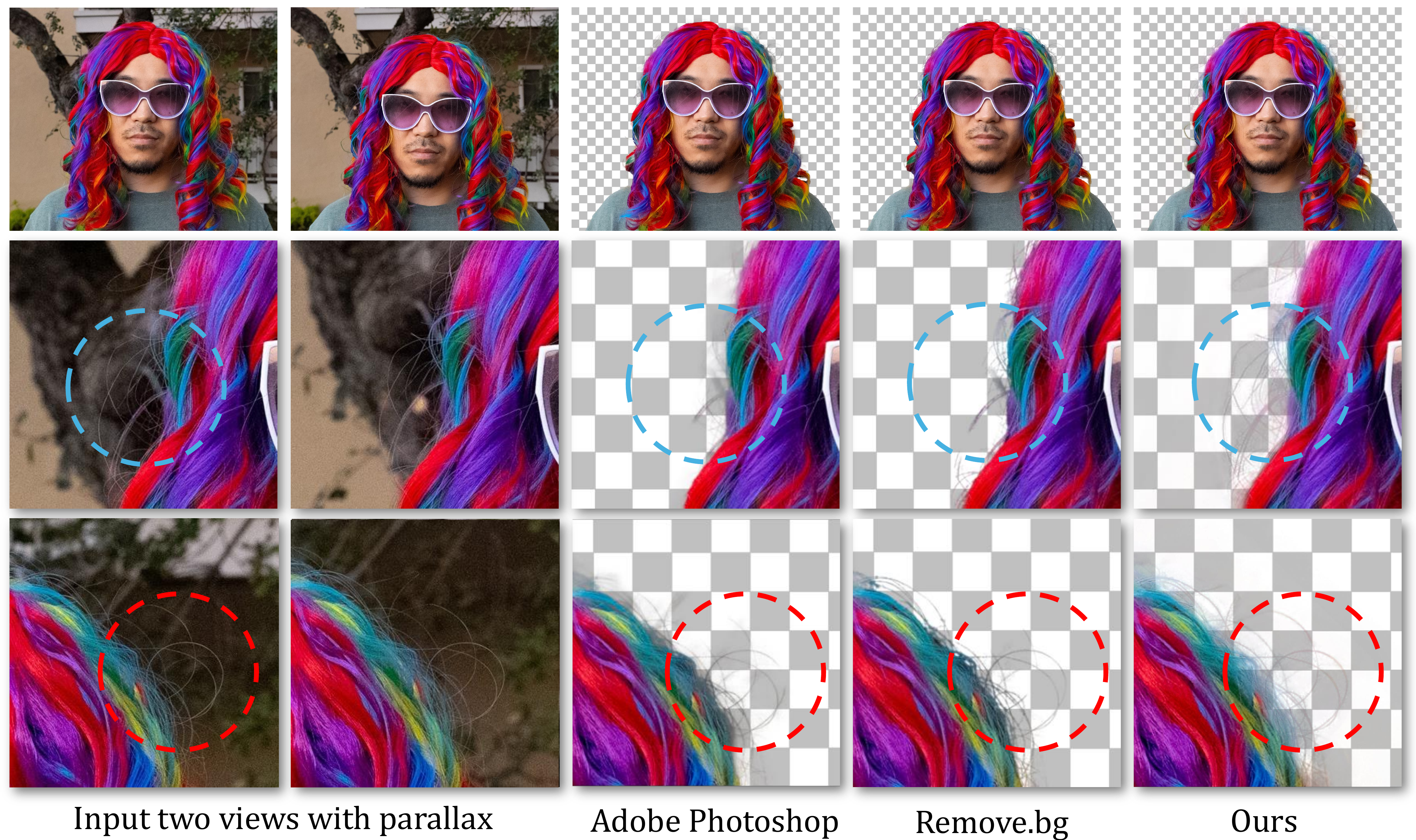}
\caption{Comparison against closed-source commercial solutions (Adobe Photoshop and Remove.bg) on real-world images. Despite competing with proprietary systems, our method recovers significantly more detail in fine structures such as hair strands.}
\label{fig:realworld_result3}
\end{figure}

We further assess the robustness of our method in real-world scenarios. For each real-world case, we capture RAW image pairs with fixed camera settings and render them consistently in Adobe Lightroom. Each scene contains slight parallax between the portrait subject and the background. Notably, our model is trained using only synthetic data and does not use any real-world matting pairs for training. Nevertheless, as shown in~\cref{fig:realworld_result}, our method consistently produces higher-quality alpha mattes and more accurate foreground colors than existing approaches. \Cref{fig:realworld_result3} further compares our method against closed-source commercial solutions (Adobe Photoshop and Remove.bg), showing that our approach recovers more detail in fine structures such as hair strands. More real-world examples and qualitative results are provided in the \textbf{supplementary material}.

\subsection{Ablation Study}
\begin{table}[htbp]
    \centering
    \caption{Ablation study on PPM-100. We evaluate five variants of our method: (1) without using the background-aligned frame; (2) without using the foreground-aligned frame; (3) single-image version (single branch); (4) same frame input to the dual-branch model; (5) adding motion noise during inference.
    }
    \label{tab:ablation}
    \small
    \setlength\tabcolsep{5pt}
    \begin{tabular}{lccc}
        \toprule
         & SAD & MSE & MSE ($\alpha F$)  \\\midrule
        Ours & \textbf{4.13} & \textbf{2.28} &  \textbf{0.55} \\
        - (1) \emph{w/o} bg-aligned frame & 7.52 & 3.78  & 1.57\\
        - (2) \emph{w/o} fg-aligned frame & 6.31 & 3.12 & 1.04\\
        - (3) \emph{w/o} 2nd frame (single branch) & 9.53 & 4.75 & 1.80 \\
        - (4) \emph{w/o} 2nd frame (same input) & 9.77 & 4.88 & 1.86 \\
        + (5) motion noise & 7.34 & 3.46 & 1.23 \\
        \bottomrule
    \end{tabular}

\end{table}

\begin{figure}[t]
  \centering
  \includegraphics[width=0.95\linewidth]{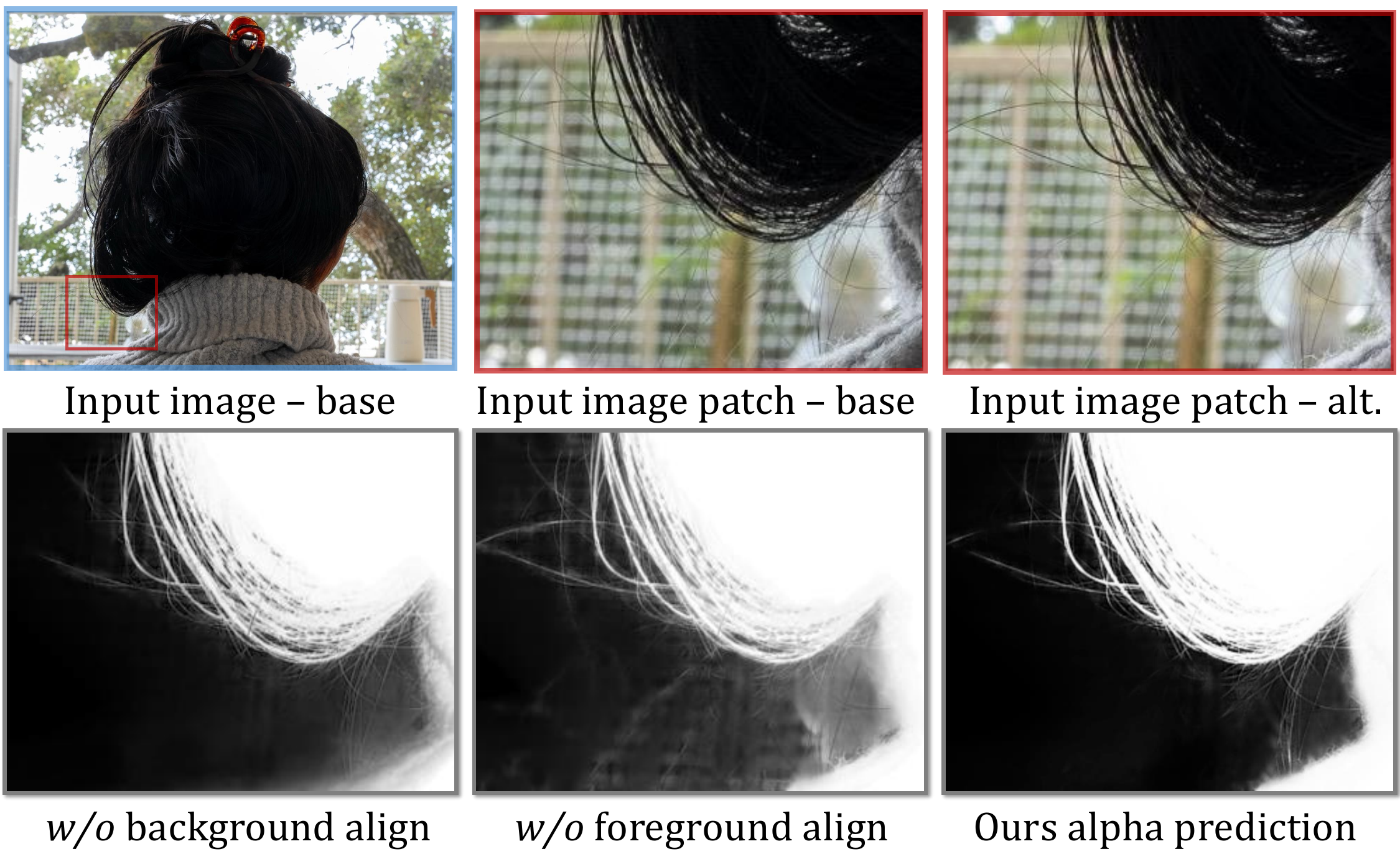}
  \caption{Ablation on real-world cases. Removing background-aligned burst loses fine details; removing foreground-aligned burst causes foreground-background confusion.}
  \label{fig:ablation}
\end{figure}

We further perform ablation studies to justify our design choices.
Removing the second branch together with both aligned inputs reduces the model to a single-image baseline, while ablating the background-aligned and foreground-aligned cues individually reveals their distinct roles.
As shown in \cref{tab:ablation}, using both cues yields the best overall performance, indicating that they provide complementary rather than redundant information.
Qualitatively, removing the background-aligned cue mainly degrades fine-detail recovery, whereas removing the foreground-aligned cue increases foreground-background confusion, as illustrated in \cref{fig:ablation}.
We also test the degenerate case where the same frame is fed into both branches; the result is close to that of the single-branch model, confirming graceful degradation when parallax is absent.
Finally, we perturb the estimated foreground flow by introducing additional motion noise. Even under this imperfect alignment, the model still outperforms the single-image baseline. This supports our central design choice: foreground alignment should not be treated as hard pixel-level evidence, but as a noisy auxiliary cue whose value is best exploited through feature-level interaction.
Taken together, these ablations indicate that the two aligned views play complementary and non-interchangeable roles, and that the consistent gains stem from our asymmetric use of the two cues rather than from the mere availability of a second frame.

\section{Conclusion \& Limitations}

We presented \emph{Parallax Portrait Matting}, showing that foreground-background parallax from slight camera motion provides a practical cue for portrait matting.
Our method uses this cue robustly by treating background alignment as reliable evidence and foreground alignment as a noisier auxiliary signal.

Our method has limitations that may point to interesting future work.
First, motion estimation is detached from training and cannot be jointly optimized.
Second, the model's performance may degrade under larger or more complex motion between frames.
Finally, parallax is not sufficient in extremely ambiguous cases, such as low-light scenes or when the subject and background remain nearly indistinguishable throughout the burst.

\section*{Acknowledgements}
This work was supported by the Research Grants Council (RGC) of Hong Kong under the Early Career Scheme (ECS) No. 24209224.

\bibliographystyle{splncs04}
\bibliography{main}

\end{document}